%% file: sigconf.tex
\documentclass[sigconf]{acmart}

\newcommand{\ours}{ SPyCE }

\usepackage{multirow}
\usepackage{colortbl}
\definecolor{BestCellRed}{RGB}{246,223,223}
\definecolor{SecondCellBlue}{RGB}{221,232,246}
\newcommand{\bestcell}[1]{\cellcolor{BestCellRed}#1}
\newcommand{\secondcell}[1]{\cellcolor{SecondCellBlue}#1}

\usepackage{pifont}
\makeatletter
\providecommand\correspondingauthor{%
\if@ACM@anonymous\else
    \g@addto@macro\addresses{\@correspondingauthormark}%
\fi}
\providecommand\@correspondingauthormark{%
\g@addto@macro\@currentauthors{%
    \advance\hfuzz by 5pt\relax\textsuperscript{\ding{41}}\relax}}
\makeatother

\AtBeginDocument{%
  }

\copyrightyear{2026}
\acmYear{2026}
\setcopyright{cc}
\setcctype{by}
\acmConference[MM '26]{Proceedings of the 34th ACM International Conference on Multimedia}{November 10--14, 2026}{Rio de Janeiro, Brazil}
\acmBooktitle{Proceedings of the 34th ACM International Conference on Multimedia (MM '26), November 10--14, 2026, Rio de Janeiro, Brazil}
\acmDOI{10.1145/3767308.3836256}
\acmISBN{979-8-4007-2213-4/2026/11}

\begin{document}


\title{SPyCE: Skill-Policy Co-evolution for Multimodal Agents}

\author{Ru Zhang}
\authornote{Equal contribution.} 
\orcid{0009-0008-7398-1512}
\affiliation{%
  \institution{Zhejiang University}
  \city{Hangzhou}
  \country{China}
}
\email{zhangru200111@gmail.com}

\author{Weijie Qiu}
\authornotemark[1]
\correspondingauthor
\orcid{0009-0006-9846-7021}
\affiliation{%
  \institution{Beijing University of Posts and Telecommunications}
  \city{Beijing}
  \country{China}
}
\email{qiuwj1001@gmail.com}



\begin{abstract}
  Multimodal agents that think with images iteratively manipulate visual evidence and invoke tools across many steps. Existing reinforcement learning methods reduce trajectories to scalar rewards, forcing the policy to discover reusable tool-use patterns from scratch on every new task; memory-based alternatives retain past experience, yet they rely on test-time retrieval, without updating the policy to absorb reusable patterns from that experience. Our key insight is that multimodal reasoning trajectories should be distilled into reusable skills that co-evolve with the policy during training, rather than being consumed as rewards or retrieved from a static store. To this end, we propose SPyCE (Skill-Policy Co-evolution), a framework that distills trajectories into a hierarchical skill library and updates it throughout reinforcement learning. Execution skills capture local visual operations, while workflow skills encode high-level priors that orchestrate tool use. During training, the policy model conditions on retrieved skills to guide its rollouts, while the skill library evolves using valuable rollouts generated by the policy. This creates a closed loop in which improved policies yield better skills, and the evolving skill library, in turn, provides stronger priors for policy rollouts. Experiments across eight benchmarks demonstrate that SPyCE consistently outperforms both RL-based and memory-based baselines. Further analysis reveals that both the hierarchical skill design and the co-evolution mechanism are critical to our design. These results suggest joint skill-policy optimization as a promising paradigm for building capable multimodal agents.
\end{abstract}

\ccsdesc[500]{Computing methodologies~Computer vision}

\keywords{Multimodal Large Language Model; Multimodal Agentic Reasoning; Hierarchical Skills}

\maketitle

\input{1_intro}

\input{2_related-work}
\input{3_method}

\input{4_exp}

\input{5_conclusion}

\begin{acks}
We extend our sincere thanks to all reviewers for their valuable efforts.
\end{acks}

\bibliographystyle{ACM-Reference-Format}
\bibliography{sample-base}










\end{document}

%% file: 1_intro.tex
\section{Introduction}
Multimodal large language models (MLLMs) have made remarkable progress in visual understanding and reasoning, largely powered by the text-based chain-of-thought (CoT) paradigm that allows models to decompose complex problems step by step ~\cite{wei2022chain,zhang2023multimodalcot,zheng2023ddcot,shao2024visual,yang2026reinforce}. Despite its effectiveness, this text-centric formulation treats visual input as a static, one-shot context: the image is consumed at the outset, and all subsequent intermediate reasoning unfolds entirely in the linguistic domain. This design introduces a fundamental  semantic gap between the richness of perceptual information and the discrete, symbolic nature of linguistic thought. Human cognition, by contrast, naturally extends beyond language: we sketch diagrams, zoom into regions of interest, and spatially rotate objects, effectively using vision as a dynamic mental scratchpad rather than a frozen snapshot. Recent advances in multimodal large language models (MLLMs) have begun to mirror this tendency, giving rise to a paradigm broadly termed
\emph{think with images}. Under this paradigm, multimodal agents iteratively invoke visual tools to actively probe, manipulate, and scrutinize visual inputs at each reasoning step~\cite{wu2023visual,yang2023mm,suris2023vipergpt,liu2024llava,su2025openthinkimg,zheng2025deepeyes,wu2025vtoolr1,guo2025thinking}.  This tool-augmented paradigm reframes tasks such as visual search, chart analysis, and interactive perception as long-horizon decision problems, where the central challenge is no longer only how to reason over pixels, but \emph{how to accumulate stable, reusable tool-use capabilities across diverse tasks}.

Reinforcement learning has become the dominant paradigm for training multimodal tool-use agents, with outcome rewards, process rewards, and tool-use rewards all demonstrating the ability to steer a policy toward more effective tool routing under  GRPO-style~\cite{shao2024deepseekmath} optimization. Yet representative methods, such as DeepEyes~\cite{zheng2025deepeyes}, OpenThinkIMG~\cite{su2025openthinkimg}, VTool-R1~\cite{wu2025vtoolr1}, and CodeVision~\cite{guo2025thinking}, share a common design assumption: a completed trajectory is valuable only as a source of scalar rewards, keeping the focus squarely on reward shaping rather than on what the trajectory itself encodes. As a consequence, recurring tool-use patterns are absorbed only implicitly through gradient updates and must be rediscovered from scratch on every new task. The root technical failure is therefore representational: successful trajectories are consumed as reward signals rather than distilled into reusable, structured knowledge, making long-horizon capability accumulation both unstable and sample-inefficient.

In the text domain, a complementary line of work improves model performance by leveraging past experience, rather than optimizing the model itself. Systems such as Reflexion~\cite{shinn2023reflexion}, Voyager~\cite{wang2023voyager}, and ExpeL~\cite{zhao2024expel} show that long-horizon behavior improves when agents can store traces, reflect on failures, and extract reusable experience. More recent methods distill trajectories into evolving cheatsheets~\cite{suzgun2026dynamic}, procedural memories~\cite{fang2025memp}, or hierarchical knowledge bases~\cite{tang2025agent} for test-time retrieval. 
While appealing, these approaches do not establish a closed feedback loop between experience and policy. Memory construction and policy improvement remain decoupled, so stronger policies do not automatically yield better experience, and improved experience does not in turn enhance subsequent policy behavior.

For multimodal tool-use agents, the challenge is greater than in text-only settings. These agents act over heterogeneous visual bottlenecks, and each tool action changes the evidence state for subsequent decisions. As a result, useful experience is not captured solely by whether a trajectory succeeds, but by the structured interaction patterns it reveals among perception, tool use, and reasoning. The flat memory structure is insufficient in this setting, because multimodal experience spans both fine-grained local execution and higher-level task organization. A single-level memory cannot clearly separate these two forms of reusable knowledge.

\vspace{-1em}
\begin{figure}[h]
\centering
\includegraphics[width=0.5\textwidth]{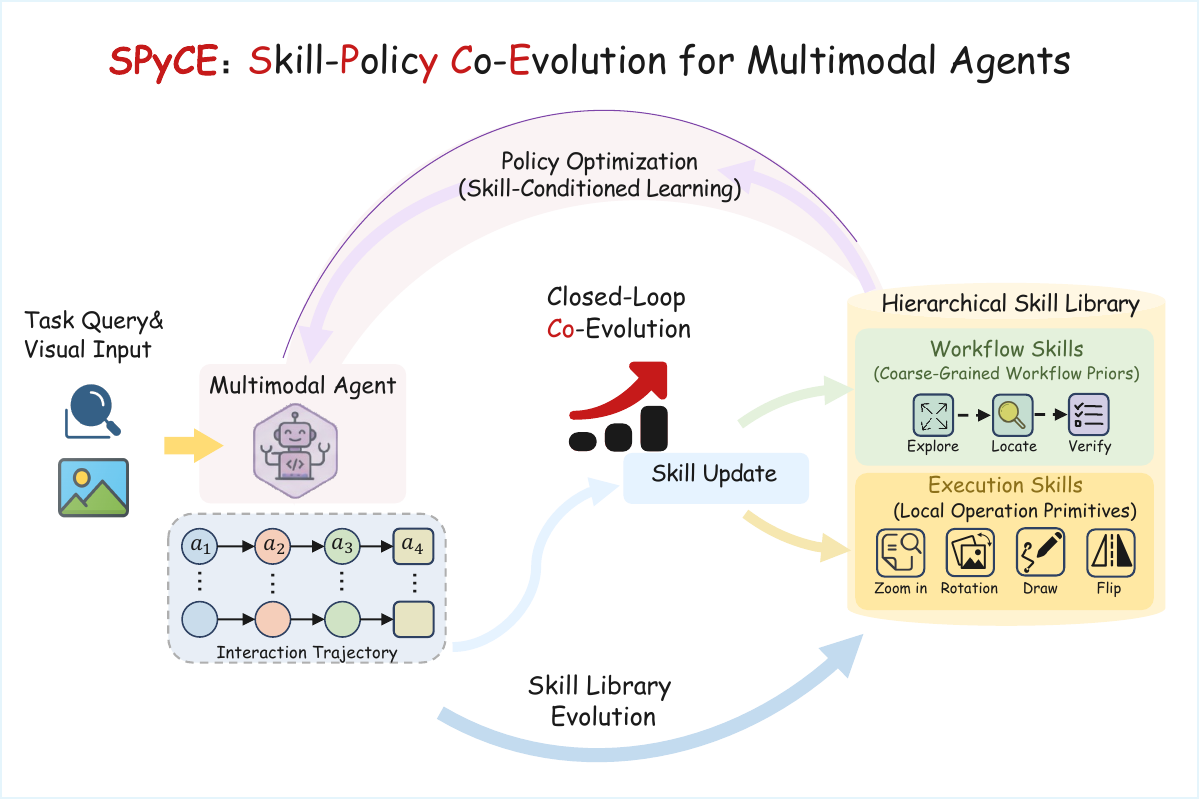}
\caption{\textbf{SPyCE} enables closed-loop co-evolution between skill and policy for multimodal agents. The agent conditions on a hierarchical skill library, including workflow skills and execution skills, to reason and act over multimodal inputs. Rollout trajectories are then distilled back into the skill library, creating a feedback loop where better skills guide better policies, and better policies induce stronger skills.}
\label{fig:teaser}
\vspace{-0.5em}
\end{figure}

To address these limitations, we propose \ours (Skill-Policy Co-evolution), a framework that closes the loop between skill and policy for multimodal tool-use agents. As illustrated in Figure~\ref{fig:teaser}, \ours distills rollout trajectories into a hierarchical skill library that co-evolves with the policy throughout reinforcement learning. The skill abstraction consists of two layers. \emph{Execution skills} encode reusable local visual operations as condition–action–effect triples, making each operation explicit in terms of when it should be applied, how it is executed, and what improvement in evidence it is expected to produce. \emph{Workflow skills} encode coarse-grained bottleneck descriptions and workflow-level priors for organizing multi-step reasoning. During execution, the policy conditions on hierarchically retrieved skills to plan and act under the current visual bottleneck. High-quality rollouts periodically refresh both skill libraries through a merge-or-add rule that suppresses noise and redundancy. This closed loop distinguishes our framework from prior work: a stronger policy produces higher-quality rollouts, better rollouts yield better skills, and better skills in turn provide stronger priors for subsequent policy updates.

We conduct comprehensive experiments on eight benchmarks including agentic reasoning, general multimodal reasoning, and visual search. The results show that \ours consistently outperforms all RL-based and memory-based baselines. Compared with RL-based methods, our framework demonstrates the importance of introducing skill abstractions into policy optimization. Compared with memory-based methods, it further highlights the advantage of co-evolving policy and skills, which enables skills not only to be accumulated but also to be continuously transformed into stronger future behavior. Ablation studies further confirm that both levels of the hierarchical skill abstraction are necessary and complementary, with workflow skills guiding global multi-step planning, execution skills supporting precise local decisions under specific visual bottlenecks, and online skill evolution further proving essential beyond a static skill library by continually refining skills from newly improved policy rollouts.

Our contributions are summarized as follows:
\begin{itemize}
  \item We identify a key limitation of existing RL-based multimodal agents: rollout trajectories are largely consumed as scalar reward signals rather than distilled into reusable knowledge. We propose SPyCE to establish an online co-evolution loop between skill and policy during training.
  \item We introduce a hierarchical skill library for multimodal tool-use agents, in which execution skills capture reusable local visual operations and workflow skills encode high-level procedural priors. Built on top of this abstraction, we develop an online co-evolution mechanism of skill and policy that periodically refreshes both skill libraries from high-quality rollouts during training.

  \item Extensive experiments on eight benchmarks demonstrate consistent and substantial gains over both RL-based and memory-based baselines. Ablation studies further reveal two essential factors behind the gains of SPyCE: the hierarchical skill design is necessary, with execution skills and workflow skills providing complementary support for local execution and global planning, while online skill-policy co-evolution is indispensable for sustaining continued policy improvement.
  
\end{itemize}

%% file: 2_related-work.tex
\section{Related Work}
\subsection{Multimodal Reasoning}
MLLMs are rapidly strengthening their vision-language reasoning capabilities. General-purpose models, including InstructBLIP~\cite{dai2023instructblip}, LLaVA~\cite{liu2023llava}, Qwen2.5-VL~\cite{bai2025qwen25vltechnicalreport}, and GPT-4V~\cite{openai2023gpt4v}, have substantially improved visual understanding and instruction following. Building on this progress, subsequent work draws inspiration from chain-of-thought reasoning in text-only domains and extends it to multimodal settings. Methods such as Visual ChatGPT~\cite{wu2023visual} and MM-REACT~\cite{yang2023mm} show that introducing explicit intermediate vision-language reasoning can yield measurable gains on challenging tasks. However, these approaches still treat images largely as static inputs: visual evidence is provided once at the beginning and then consumed passively during inference, rather than being actively revisited, updated, or manipulated as reasoning unfolds. Recent work extends this paradigm toward multimodal tool-use agents by enabling models to interact with external tools over dynamically evolving visual evidence. OpenAI o3 \cite{openai2025o3o4mini} marks an important step in this direction, as it can incorporate images directly into the reasoning process, a capability referred to as \emph{thinking with images}. 

More recent systems such as DeepEyes~\cite{zheng2025deepeyes}, OpenThinkIMG~\cite{su2025openthinkimg}, VTool-R1~\cite{wu2025vtoolr1}, and CodeVision~\cite{guo2025thinking} push this paradigm further by explicitly training policy models to reason with images and use tools across multi-step interactions. However, these methods primarily emphasize improved reward design, while largely overlooking the intrinsic value of interaction trajectories. Rather than being distilled into reusable structured knowledge, trajectories are mostly reduced to scalar signals for policy optimization. By contrast, our framework makes fuller use of interaction trajectories by distilling them into skills and enabling the co-evolution of skills and policy.

\subsection{Experience Abstraction and Agent Skill Learning}
Empowering agents to learn from past interactions has become a promising direction for further improving agent capabilities. Representative systems such as Generative Agents~\cite{park2023generative}, Reflexion~\cite{shinn2023reflexion}, Voyager~\cite{wang2023voyager}, and ExpeL~\cite{zhao2024expel} show that long-horizon behavior can benefit from retaining interaction histories, reflecting on failures, and extracting reusable experience.  More recent methods make this design more explicit: Dynamic Cheatsheet~\cite{suzgun2026dynamic} maintains an evolving cheatsheet for test-time reuse, MemP~\cite{fang2025memp} models procedural memory over multi-step interaction, and Agent KB~\cite{tang2025agent} organizes cross-domain experience into a hierarchical repository. These methods suggest an alternative path for agent evolution beyond parameter updates. Multimodal agents have likewise begun to abstract trajectories into structured skills. Mirage-1~\cite{xie2025mirage} and XSkill~\cite{jiang2026xskill} organize experience into hierarchical or task- and action-level skills grounded in visual observations; MMSkills~\cite{zhang2026mmskills} augments procedures with state cards and keyframes. A parallel line couples skills with reinforcement learning: SAGE~\cite{wang2025sage} and SkillRL~\cite{xia2026skillrl} integrate evolving skill libraries into policy optimization, ReSkill~\cite{he2026reskill} revises skills alongside the policy. Our work targets multimodal reasoning under heterogeneous visual observations and closes the loop between skills and policy through joint co-evolution during training.



%% file: 3_method.tex
\section{Methodology}

\begin{figure*}[t]
\centering
\includegraphics[width=\textwidth]{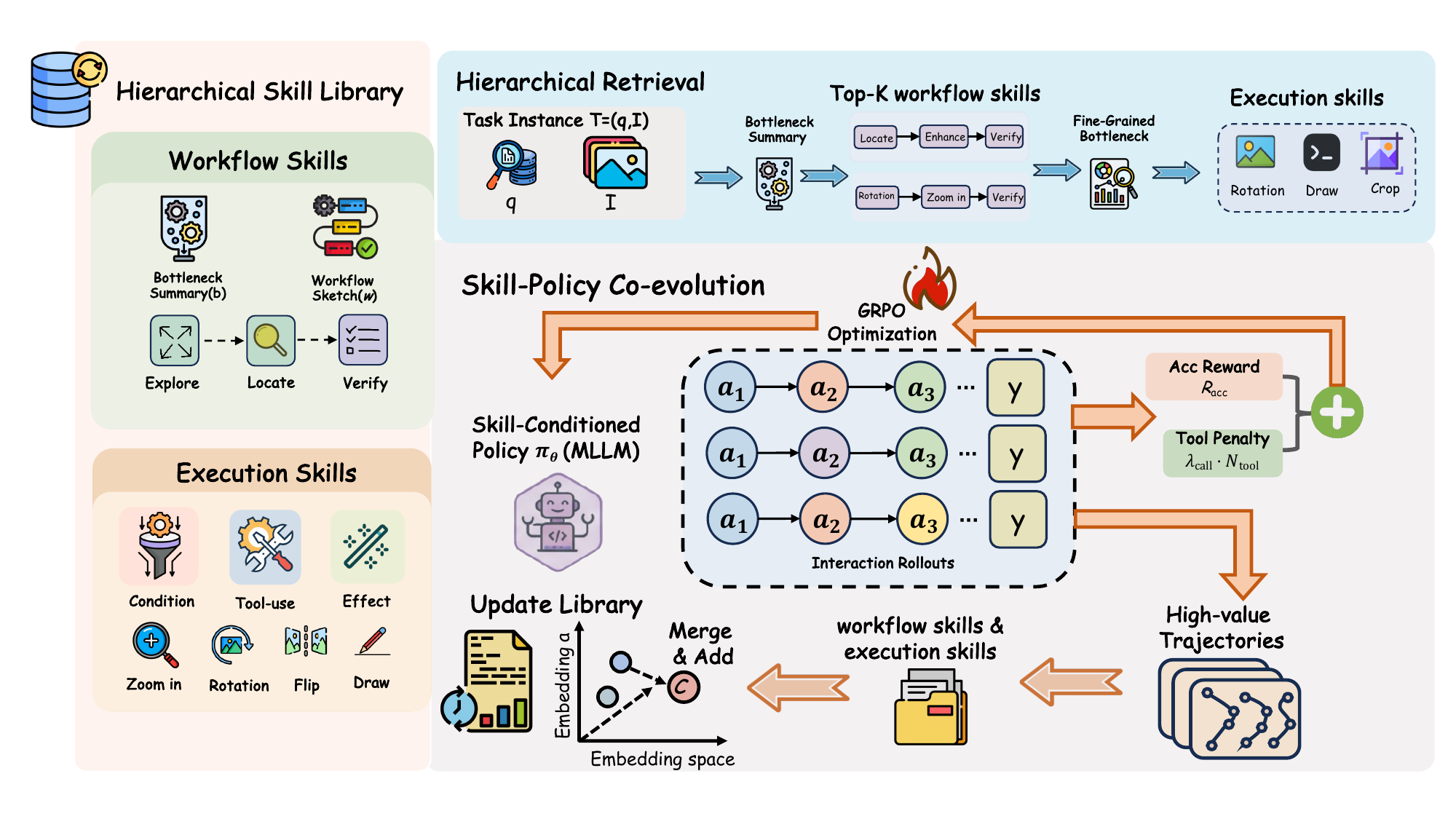}
\vspace{-3em}
\caption{Overview of the proposed skill-policy co-evolution framework for multimodal tool-use reasoning. First, we collect interaction trajectories and distill them into a hierarchical skill library. Then the policy performs skill-conditioned reasoning by retrieving high-level workflow priors and low-level executable skills, while the skill  library is updated from high-quality rollouts. This forms a closed loop between policy improvement and skill evolution.}
\label{fig:pipeline}
\end{figure*}

We consider multimodal visual reasoning tasks in which an MLLM grounds its reasoning in task-relevant visual evidence by actively acting on and interacting with images through tool use. We first formalize the task setting and present the optimization objective that underlies our framework (Section~\ref{sec:formulation}). 
As shown in Figure~\ref{fig:pipeline}, our framework couples policy optimization with a hierarchical skill library through two stages: (1) hierarchical skill library construction from successful trajectories during policy evolution (Section~\ref{sec:skill_library}); and (2) a skill-policy co-evolution stage, in which retrieved skills guide policy optimization while improved rollouts are continuously distilled back into the skill library (Section~\ref{sec:skill_evolution}).

\subsection{Preliminaries}
\label{sec:formulation}
\noindent \textbf{Problem Formulation.} We formulate multimodal tool-use reasoning as a partially observable Markov decision process (POMDP)~\cite{kaelbling1998planning}. A task instance $T=(q,\mathcal{I})$ consists of a natural language query $q$ and an image set $\mathcal{I}=\{I_1,\dots,I_m\}$. At each step $t$, the agent receives an observation $o_t$ and selects a tool action $a_t$, which produces a new observation and updates the interaction state. This process yields a trajectory $\tau=[(s_0,a_0,o_0),\dots,(s_T,a_T,o_T)]$, after which the agent outputs a final answer $\hat{y}$.

\noindent \textbf{Group-based Policy Optimization.}  We optimize the policy with reinforcement learning by maximizing the expected reward of sampled tool-use trajectories. For each task input $x_i$, we sample a group of $N$ rollouts $\{\tau_i^{(n)}\}_{n=1}^{N} \sim \pi_{\theta_{\mathrm{old}}}(\cdot \mid x_i)$ from the current policy. Each rollout $\tau_i^{(n)}$ is assigned a scalar reward $R(\tau_i^{(n)})$, and policy optimization aims to increase the likelihood of rollouts with higher rewards. Two representative instances are GRPO \cite{shao2024deepseekmath} and RLOO~\cite{ahmadian2024back}. GRPO normalizes each rollout reward within its sampled group:
$$
A_i =
\frac{R_i - \mathrm{mean}(\{R_j\}_{j=1}^{N})}
{\mathrm{std}(\{R_j\}_{j=1}^{N}) + \varepsilon},
$$
where $R_i = R(\tau_i)$. The corresponding clipped GRPO objective is
\[
\begin{aligned}
\mathcal{L}_{\mathrm{GRPO}}(\theta)
=
\mathbb{E}_i \Big[
&\min\!\Big(
r_i(\theta)A_i,\,
\mathrm{clip}\!\big(r_i(\theta),1-\epsilon,1+\epsilon\big)A_i
\Big) \\
&-\beta D_{\mathrm{KL}}\!\big(\pi_\theta(\cdot\mid x_i)\,\|\,\pi_{\mathrm{ref}}(\cdot\mid x_i)\big)
\Big].
\end{aligned}
\]

where $ r_i(\theta)=\frac{\pi_\theta(\tau_i\mid x_i)}{\pi_{\theta_{\mathrm{old}}}(\tau_i\mid x_i)} $ is the importance ratio, and $\pi_{\mathrm{ref}}$ is a reference policy used for KL regularization. RLOO differs from GRPO by replacing the normalized group statistic with a leave-one-out baseline: $ A_i^{\mathrm{RLOO}} = R_i - \frac{1}{N-1}\sum_{j \neq i} R_j. $

In our method, the reward for each rollout is defined as
$$
R(\tau)=R_{\mathrm{acc}}(\tau)-\lambda_{\mathrm{call}}\cdot \mathbf{1}\!\left[\mathrm{Succ}(\tau)=1\right]\cdot N_{\mathrm{tool}}(\tau),
$$
where $R_{\mathrm{acc}}(\tau)=r(\tau)\in\{0,1\}$ indicates task success, $\mathrm{Succ}(\tau)$ indicates whether the task is solved successfully, and $N_{\mathrm{tool}}(\tau)$ denotes the number of tool calls. Applying the tool-use penalty only to successful rollouts discourages redundant operations without unnecessarily suppressing exploration in failed ones.

\subsection{Hierarchical Skill Library}
\label{sec:skill_library}

During the first stage of policy optimization, we construct the skill library from interaction trajectories to provide reusable guidance for subsequent learning. Multimodal tool-use reasoning requires skills at two complementary granularities: workflow skills for high-level task guidance and execution skills for fine-grained decision support during interaction. Local visual bottlenecks, such as rotated evidence, tiny targets, or cluttered regions, call for executable short-horizon operations, while more complex tasks also depend on high-level workflow priors for tool orchestration and multi-step reasoning. Therefore, we hierarchically organize the skill space into two levels: an execution-skill layer $\mathcal{K}$ and a workflow-skill layer $\mathcal{H}$.

\subsubsection{Execution Skills.} 
Execution skills encode reusable operations for resolving local visual bottlenecks. Each execution skill is represented as a tuple \textbf{$k=(c,u,e)$}, where $c$ denotes a textual trigger condition specifying when the skill applies, $u$ denotes the tool action or action template to be used, and $e$ denotes the expected local effect. This representation makes each skill explicit about when it applies, how it should be used, and what local evidence improvement it is expected to produce. To distill execution skills, we employ a specialized MLLM named $\textsc{MLLM}_{\mathrm{dk}}$ for skill-library operations. Given a successful trajectory $\tau$ and its associated task input $(q,I)$, we prompt $\textsc{MLLM}_{\mathrm{dk}}$ to summarize reusable local tool-use patterns into execution-skill tuples:
$$
\{k_j\}_{j=1}^{K_\tau}=\textsc{MLLM}_{\mathrm{dk}}(q,I,\tau),
$$
where $K_\tau$ is the number of execution skills distilled from $\tau$. Each $k_j$ is a compact textual abstraction of a reusable local pattern, rather than a raw trajectory fragment. For example, a successful trajectory may reveal that rotated evidence can be resolved by image rotation, while small or boundary-adjacent evidence can be addressed by local cropping. Because execution skills are represented as condition--action--effect tuples, they can be directly matched and reused based on visual conditions, without requiring similarity to an entire trajectory.

\subsubsection{Workflow Skills.}
We introduce \textit{workflow skills} to capture workflow-level priors for multi-step tool orchestration and reasoning. Each workflow skill is represented as \textbf{$h=(b,w),$}
where $b$ denotes a coarse-grained summary of the visual bottleneck (e.g., cross-region consistency checking), and $w$ denotes a workflow sketch that summarizes the corresponding high-level solution pattern (e.g., locate evidence $\to$ enhance local content $\to$ verify consistency). Unlike execution skills, workflow skills do not directly encode executable actions. Instead, they distill workflow-level patterns for orchestrating tool use from successful trajectories. Consitent with the execution skills, workflow skills are distilled with $\textsc{MLLM}_{\mathrm{dk}}$ from the task query $q$, the image input $I$, and a successful trajectory $\tau$:
$$
h=(b,w)=\textsc{MLLM}_{\mathrm{dk}}(q,I,\tau),
$$
where $b$ denotes the bottleneck description and $w$ denotes the corresponding workflow sketch. This distillation step converts successful trajectories into compact workflow-level abstractions that are transferable across tasks and suitable for retrieval.

\subsubsection{Skill Library Consolidation.}
Without consolidation, newly distilled skills would be inserted into the libraries directly, causing two issues: increased retrieval noise from near-duplicate entries and uncontrolled library growth due to entry proliferation. To address this, we apply a unified merge-or-add rule to both libraries.
For each newly distilled skill candidate, the system checks whether any existing entry is sufficiently similar in the embedding space. For execution skills, similarity is computed based on the trigger condition. Given a candidate $\hat{k}=(\hat{c},\hat{u},\hat{e})$, the system checks whether any existing execution skill $k=(c,u,e)\in\mathcal{K}$ has cosine similarity above a threshold $\theta_{\mathrm{sim}}^{K}$ with $\hat{c}$. If such an entry exists, the candidate is merged into the most similar matched skill by refining the items; otherwise, it is added as a new entry. For workflow skills, matching is performed using the bottleneck description. Given a candidate $\hat{h}=(\hat{b},\hat{w})$, the system checks whether any existing workflow skill $h=(b,w)\in\mathcal{H}$ has cosine similarity above $\theta_{\mathrm{sim}}^{H}$ with $\hat{b}$. If so, the candidate is merged into the matched entry; otherwise, it is inserted as a new workflow skill. To further control library size, we record the number of similarity-based matches for each entry and prune the least frequently matched ones whenever the library exceeds its capacity.

\subsection{Skill and Policy Co-Evolution}
\label{sec:skill_evolution}
During this stage, policy optimization is performed on skill-conditioned rollouts at every training step, whereas the skill libraries evolve periodically every $N$ steps. This design enables a closed loop of co-evolution between the policy and the skill library: retrieved skills provide structured guidance for policy improvement, and the improved policy in turn produces higher-quality rollouts from which better skills can be distilled.

\subsubsection{Hierarchical Skill Retrieval.}
We adopt a two-stage retrieval procedure aligned with the hierarchy of workflow skills and execution skills. The key idea is to first retrieve high-level workflow priors, and then retrieve task-relevant execution skills conditioned on the identified bottleneck. Given a task $T=(q,I)$, the actor model first produces a coarse bottleneck query $ \tilde{b}_{T}={MLLM}_(q,I).$ This query is then used to retrieve the top-$K$ workflow skills whose bottleneck descriptions are most similar in the embedding space:
$\mathcal{H}_{T}^{\mathrm{top}} = \operatorname{TopK}_{h=(b,w)\in\mathcal{H}} \;\mathrm{sim}(\tilde{b}_{T}, b).$ 
The actor model then further decomposes the task into fine-grained local bottleneck descriptors conditioned on the retrieved workflow priors:
\[
\mathcal{B}=\{b^{(1)}, \dots, b^{(M)}\}
= \mathrm{\pi_{\theta}}(q,I,\mathcal{H}_{T}^{\mathrm{top}}).
\]
Each descriptor $b^{(m)}$ specifies a local visual obstacle, such as viewpoint mismatch, a small target, or dense text. For each descriptor, we retrieve the best-matching execution skill by comparing it with the trigger conditions of skills in the execution-skill library $\mathcal{K}$:
\[
k^{*(m)}
=
\arg\max_{k=(c,u,e)\in\mathcal{K}}\;
\cos\!\big(E_{L}(b^{(m)}),\, E_{L}(c)\big).
\]
If the best similarity score falls below a threshold $\theta_{sim}$, the descriptor is discarded. The retained set of execution skills is
\[
\mathcal{K}_{T}^{L}
=
\left\{
k^{*(m)}
\;\middle|\;
\max_{k\in\mathcal{K}}\, S_{L}(k\mid b^{(m)})\ge\tau_{sim},\;
b^{(m)}\in\mathcal{B}
\right\}.
\]

The policy then selects actions conditioned on both the retrieved workflow skills and the retrieved execution skills:
\[
a_t \sim \pi_{\theta}\!\left(a_t \mid o_{\le t},\, q,\, I,\, \mathcal{H}_{T}^{\mathrm{top_{1}}},\, \mathcal{K}_{T}^{L}\right),
\]
where $o_{\le t}=(o_0,\dots,o_t)$ denotes the observation history. The resulting skill-conditioned trajectories are scored using the reward defined in Section~\ref{sec:formulation} and optimized with the GRPO objective. This hierarchical retrieval guides rollout generation with reusable workflow-level and execution-level priors, rather than leaving policy updates to rely on generic task interactions alone. As a result, it improves both the efficiency and the stability of policy optimization.

\subsubsection{Online Skill Evolution.}
Skills induced from early, weaker rollouts may become suboptimal as the policy improves. We therefore perform periodic skill library evolution using newly collected high-quality rollouts. After every $N$ policy steps, we open an evolution window indexed by $e$. Rather than updating the libraries from all successful trajectories, we retain only the top trajectories ranked by GRPO relative advantage $A(\tau)$, which preserves diverse but high-quality interaction patterns for subsequent skill evolution. For workflow skills, let $\mathcal{T}_{e}^{\mathrm{wf}}$ denote the retained top-$\rho_{\mathrm{wf}}$ successful rollouts in window $e$ for task $T$. We distill a new workflow-skill candidate with $\textsc{MLLM}_{\mathrm{dk}}$ as
\[
\hat{h}_{e}=\bigl(\hat{b}_{e},\,\hat{w}_{e}\bigr)
= \textsc{MLLM}_{\mathrm{dk}}\!\left(q,I,\mathcal{T}_{e}^{\mathrm{wf}}\right),
\]
where $\hat{b}_{e}$ is the updated bottleneck description and $\hat{w}_{e}$ is the corresponding workflow sketch. The candidate is then merged into, or added to, the workflow-skill library $\mathcal{H}$ using the consolidation rule in Section~\ref{sec:skill_library}. Execution skills are distilled in the same manner from the retained top-$\rho_{\mathrm{ex}}$ successful rollouts for execution-skill evolution. To control library growth, we further prune the least useful entries of execution skills when the library exceeds its capacity. In particular, we rank execution skills by their empirical success rate and remove the lowest-ranked ones, where the empirical success rate of an execution skill $k$ is defined as $ \mathrm{SR}(k)=\frac{N_{\mathrm{success}}(k)}{\max\!\left(N_{\mathrm{retrieval}}(k),1\right)}.$ This periodic evolution closes the loop: stronger policy produces higher-quality rollouts, which yield better skills, which in turn provide stronger priors for subsequent rollouts. The co-evolution mechanism is what distinguishes our method from static memory systems that only retrieve at test time.

%% file: 4_exp.tex
\section{Experiments}

\subsection{Experimental Setup}

\subsubsection{Tasks and Data.} Following CodeVision \cite{guo2025thinking}, we employ a two-stage post-training recipe, with supervised fine-tuning (SFT) followed by reinforcement learning (RL). In the SFT stage, we use all publicly available CodeVision-SFT data (about 5K samples). In the RL stage, we start from the released 34,795 CodeVision-RL samples and refine them through several filtering steps, including tool-demand screening, tool-type balancing, duplicate removal, and rollout-cost control, resulting in a final set of about 12K samples. We run evaluation across three multimodal domains: (1) TIR-Bench \cite{li2025tir} for agentic thinking with images; (2) MathVerse \cite{zhang2024mathverse}, MathVision \cite{wang2024measuring}, WeMath \cite{qiao2025we} and ChartQAPro \cite{masry2025chartqapro} for multimodal reasoning; (3) V* Bench \cite{wu2024v}, HRBench-4K \cite{wang2025divide}, and HRBench-8K \cite{wang2025divide} for visual search.

\subsubsection{Baselines and Metrics.}
We compare our method against baselines from three categories: 
(1) \textbf{prompt-based methods}, including Prompt with and without Tool Examples; 
(2) \textbf{RL-based methods}, including GRPO \cite{shao2024deepseekmath} and RLOO~\cite{ahmadian2024back}; and 
(3) \textbf{memory-based methods}, including MemP\cite{fang2025memp}, Dynamic Cheatsheet~\cite{suzgun2026dynamic}, and Agent-KB~\cite{tang2025agent}. The details regarding baseline implementations can be found in the supplementary materials. For multimodal reasoning and visual search benchmarks, we report the \textbf{Task Success Rate}. For agentic thinking with images, we report \textbf{Task Success Rate} and \textbf{Avg. Tool Calls} to evaluate effectiveness and efficiency, respectively.

\subsubsection{Implementation.}  Our method is trained on Qwen3-VL-4B-Instruct\cite{bai2025qwen3} and Qwen3-VL-8B-Instruct\cite{bai2025qwen3} as the backbone models. For skill distillation, we employ Qwen3-VL-235B-A22B-Instruct ~\cite{bai2025qwen3} as $\textsc{MLLM}_{\mathrm{dk}}$, and for retrieval, we use text-embedding-3-small~\cite{openaiembeddings3}. For the SFT stage, we train for 2 epochs with a batch size of 128, a learning rate of 5e-6, a cosine learning rate scheduler, and a warmup ratio of 0.05. For the RL stage, we first train one epoch for skill library initialization, and then conduct one epoch with skill-conditioned optimization. We set the learning rate to 1e-6 and the batch size to 64, and perform 8 rollouts per sample. The coefficients for the tool-use penalty is set to 0.05. For evaluation on the benchmark, we retain only skill retrieval and remove all other components, in order to provide a clean assessment of the effectiveness of our framework. For more detailed information on training and evaluation settings, please see the supplementary material.

\subsection{Main Results}

\begin{table*}[htbp]
\caption{Overall accuracy on multimodal reasoning and visual search benchmarks. The best results are highlighted in red, and the second-best results are highlighted in blue.}
\centering
\footnotesize
\setlength{\tabcolsep}{4pt}
\renewcommand{\arraystretch}{0.98}
\begin{tabular}{l|ccc|cccc}
\toprule
\textbf{Method} & \textbf{V*} & \textbf{HRBench-4K} & \textbf{HRBench-8K} & \textbf{ChartQAPro} & \textbf{MathVerse} & \textbf{MathVision} & \textbf{WeMath} \\
\midrule

\multicolumn{8}{c}{\textbf{Qwen3-VL-4B-Instruct}} \\
\addlinespace[1pt]
\multicolumn{8}{c}{\textit{Prompt-based Methods}} \\
\cmidrule(lr){1-8}
Prompt w/o Tool Examples & 73.8 & 71.6 & 66.8 & 45.6 & 46.8 & 51.6 & 53.9 \\
Prompt w/ Tool Examples  & 74.3 & 72.1 & 67.8 & 46.4 & 47.9 & 52.2 & 54.8 \\
\midrule
\multicolumn{8}{c}{\textit{RL-based Methods}} \\
\cmidrule(lr){1-8}
GRPO \cite{shao2024deepseekmath}
& \secondcell{78.5} & \secondcell{76.0} & \secondcell{70.4} & 49.5 & \secondcell{52.9} & 55.3 & 56.1 \\
RLOO~\cite{ahmadian2024back}
& 77.5 & 74.8 & 69.4 & 49.1 & 51.6 & \secondcell{55.9} & \secondcell{57.3} \\
\midrule
\multicolumn{8}{c}{\textit{Memory-based Methods}} \\
\cmidrule(lr){1-8}
MemP~\cite{fang2025memp}
& 75.4 & 73.4 & 67.6 & 48.6 & 47.2 & 53.3 & 56.7 \\
Dynamic Cheatsheet~\cite{suzgun2026dynamic}
& 74.9 & 74.1 & 67.5 & 48.6 & 48.8 & 54.0 & 56.2 \\
Agent-KB~\cite{tang2025agent}
& 76.4 & 74.8 & 68.4 & \secondcell{50.4} & 49.8 & 54.8 & 57.1 \\
\midrule
Ours
& \bestcell{80.6} & \bestcell{78.8} & \bestcell{73.9} & \bestcell{51.4} & \bestcell{55.0} & \bestcell{57.4} & \bestcell{59.7} \\
\midrule

\multicolumn{8}{c}{\textbf{Qwen3-VL-8B-Instruct}} \\
\addlinespace[1pt]
\multicolumn{8}{c}{\textit{Prompt-based Methods}} \\
\cmidrule(lr){1-8}
Prompt w/o Tool Examples & 77.5 & 72.2 & 68.4 & 49.2 & 62.2 & 53.9 & 62.5 \\
Prompt w/ Tool Examples  & 78.0 & 73.1 & 68.4 & 49.7 & 62.9 & 54.6 & 63.0 \\
\midrule
\multicolumn{8}{c}{\textit{RL-based Methods}} \\
\cmidrule(lr){1-8}
GRPO \cite{shao2024deepseekmath}
& \secondcell{82.2} & \secondcell{77.2} & \secondcell{72.9} & 50.7 & \secondcell{66.5} & 57.7 & \secondcell{66.6} \\
RLOO~\cite{ahmadian2024back}
& 81.2 & 76.1 & 70.8 & \secondcell{51.6} & 65.8 & \secondcell{58.1} & 65.8 \\
\midrule
\multicolumn{8}{c}{\textit{Memory-based Methods}} \\
\cmidrule(lr){1-8}
MemP~\cite{fang2025memp}
& 80.1 & 73.6 & 69.4 & 49.6 & 64.7 & 55.8 & 64.4 \\
Dynamic Cheatsheet~\cite{suzgun2026dynamic}
& 79.1 & 74.8 & 69.6 & 49.7 & 64.9 & 56.1 & 64.6 \\
Agent-KB~\cite{tang2025agent}
& 80.6 & 75.2 & 70.4 & 51.1 & 65.7 & 56.6 & 65.8 \\
\midrule
Ours
& \bestcell{84.3} & \bestcell{78.0} & \bestcell{74.3} & \bestcell{52.5} & \bestcell{68.7} & \bestcell{61.2} & \bestcell{67.8} \\
\bottomrule
\end{tabular}
\label{tab:main_results}
\end{table*}

\begin{table*}[htbp]
\caption{Category-level accuracy (\%) on TIR-Bench under the \textbf{Qwen3-VL-8B-Instruct} backbone. 
\textbf{All} denotes the overall average accuracy across all tasks, and \textbf{\# Calls} denotes the average number of tool calls. The best results are highlighted in red, and the second-best results are highlighted in blue. For \textbf{\# Calls}, lower is better.}
\centering
\scriptsize
\renewcommand{\arraystretch}{1.12}
\resizebox{\textwidth}{!}{
\begin{tabular}{l c c c c c c c c c c c c c c c}
\toprule
\textbf{Method} 
& \textbf{All}
& \textbf{Color}
& \textbf{Pro.}
& \textbf{OCR}
& \textbf{SR}
& \textbf{Maze}
& \textbf{Math}
& \textbf{WS}
& \textbf{LL-VQA}
& \textbf{IR}
& \textbf{SD}
& \textbf{JG}
& \textbf{VS}
& \textbf{RG}
& \textbf{\# Calls} \\
\midrule

\multicolumn{16}{c}{\textit{Prompt-based Methods}} \\
\cmidrule(lr){1-16}
Prompt w/o Tool Examples
& 21.9 & 25.0 & 26.7 & 46.7 & 18.0 & 20.8 & 30.8 & 2.0 & 30.0 & 16.3 & 21.0 & 5.8 & 35.0 & 13.3 & / \\
Prompt w/ Tool Examples
& 22.9 & 26.0 & 28.3 & 50.0 & 16.0 & 20.8 & 34.2 & 3.0 & \secondcell{34.0} & 17.5 & 21.0 & 6.7 & 34.2 & 13.3 & \bestcell{0.86} \\
\midrule

\multicolumn{16}{c}{\textit{RL-based Methods}} \\
\cmidrule(lr){1-16}
GRPO \cite{shao2024deepseekmath}
& \secondcell{27.7} & 26.0 & 28.3 & \secondcell{53.3} & \bestcell{30.0} & \secondcell{21.7} & \secondcell{37.5} & 5.0 & \bestcell{36.0} & \secondcell{21.3} & 23.0 & \secondcell{10.8} & \secondcell{40.8} & \secondcell{44.0} & 5.13 \\
RLOO~\cite{ahmadian2024back}
& 26.7 & 24.0 & \secondcell{29.2} & 51.7 & \secondcell{28.0} & 20.8 & 35.8 & \bestcell{7.0} & 30.0 & \secondcell{21.3} & \secondcell{25.0} & 9.2 & 38.3 & 42.7 & 5.38 \\
\midrule

\multicolumn{16}{c}{\textit{Memory-based Methods}} \\
\cmidrule(lr){1-16}
MemP~\cite{fang2025memp}
& 23.4 & 25.0 & 28.3 & 46.7 & 18.0 & 20.0 & 34.2 & 1.0 & 32.0 & 17.5 & 20.0 & 6.7 & 37.5 & 25.3 & 5.39 \\
Dynamic Cheatsheet~\cite{suzgun2026dynamic}
& 23.5 & \secondcell{28.0} & 28.3 & 50.0 & 14.0 & 18.3 & 32.5 & 1.0 & 28.0 & \secondcell{21.3} & 21.0 & 10.0 & 33.3 & 28.0 & 5.03 \\
Agent-KB~\cite{tang2025agent}
& 24.8 & 27.0 & \secondcell{29.2} & 48.3 & 22.0 & 18.3 & 33.3 & 4.0 & 30.0 & \secondcell{21.3} & 22.0 & 9.2 & 36.7 & 32.0 & 6.71 \\
\midrule

Ours
& \bestcell{32.0} & \bestcell{30.0} & \bestcell{30.0} & \bestcell{58.3} & \secondcell{28.0} & \bestcell{26.7} & \bestcell{44.2} & \secondcell{6.0} & \bestcell{36.0} & \bestcell{31.3} & \bestcell{31.0} & \bestcell{15.0} & \bestcell{43.3} & \bestcell{52.0} & \secondcell{4.83} \\
\bottomrule
\end{tabular}
}
\label{tab:tir_results}
\end{table*}

The results in Table~\ref{tab:main_results} and Table~\ref{tab:tir_results} show that our method achieves the best performance across all benchmarks. 
Compared with prompting alone, the gains are substantial, especially on interaction-heavy tasks such as V* and HRBench, showing that simple in-context tool demonstrations are insufficient to fully unlock effective multimodal problem solving. Our method also remains consistently stronger than vanilla RL baselines such as GRPO and RLOO, indicating that the improvement does not merely come from policy optimization, but from the structured skill guidance introduced by our framework. In addition, although memory-based methods provide moderate improvements over prompting, they still lag behind our method, suggesting that naive experience retrieval is less effective than explicitly organizing and reusing transferable skills. The category-level results on TIR-Bench provide a more detailed view of where the gains arise. Our method achieves the best overall score (\textbf{All} = 32.0) while using fewer tool calls than all RL-based and memory-based baselines, indicating a more favorable effectiveness--efficiency trade-off. The improvements are concentrated in categories that involve sequential transformations and nontrivial tool orchestration, including OCR, Maze, Math, IR (Instrument Reasoning), SD (Spot Difference), JG (Jigsaw Game), VS (Visual Search), and RG (Rotation Game). These tasks require the agent to not only invoke tools, but also decide when to apply them, how to chain them, and how to exploit intermediate outputs. This pattern is consistent with the design of skill-policy co-evolution: the evolving skill library provides structured guidance for policy learning, while the progressively improved policy, in turn, facilitates the extraction of more advanced skills from interaction trajectories.

\subsection{Ablation Studies}

\begin{table}[htbp]
\centering
\small
\caption{Ablation results for Qwen3-VL-8B-Instruct on TIR-Bench. We report average accuracy and average tool calls. For ablated variants, the value in parentheses indicates the absolute accuracy drop with respect to the full model.}
\label{tab:ablation}
\begin{tabular}{lcc}
\hline
\textbf{Method} & \textbf{Acc.} & \textbf{Tool Calls} \\
\hline
Full Model & \bestcell{32.0} & \secondcell{4.83} \\
\hline
\multicolumn{3}{l}{\textit{Skill Structure}} \\
\quad w/o Workflow Skills & 28.0 \textcolor{red}{\small(4.0$\downarrow$)} & 5.91 \\
\quad w/o Execution Skills & 28.7 \textcolor{red}{\small(3.3$\downarrow$)} & 5.33 \\
\hline
\multicolumn{3}{l}{\textit{Skill Evolution}} \\
\quad Frozen Skills & \secondcell{28.9} \textcolor{red}{\small(3.1$\downarrow$)} & 5.54 \\
\hline
\multicolumn{3}{l}{\textit{Reward Design}} \\
\quad w/o Tool-call Penalty & 29.4 \textcolor{red}{\small(2.6$\downarrow$)} & 6.83 \\
\quad Always Penalize Tool Calls & 26.1 \textcolor{red}{\small(5.9$\downarrow$)} & \bestcell{3.96} \\
\hline
\end{tabular}
\end{table}

We conduct ablation experiments to quantify the contribution of each component, with the results reported in Table~\ref{tab:ablation}. We highlight three findings.
(1) \textbf{Skill Structure.} Both levels of the hierarchical skill structure are important. Removing high-level workflow skills reduces the average accuracy by 4.0\% and increases the average number of tool calls from 4.83 to 5.91, indicating that high-level skills play a central role in global planning and in coordinating multi-step tool-use workflows. Removing low-level execution skills also leads to a clear degradation, with accuracy dropping to 28.7\% and tool calls increasing to 5.33. These results show that the two skill levels are complementary: high-level skills provide coarse-grained procedural guidance, while low-level skills support accurate local execution.
(2) \textbf{Skill Evolution.} Online skill evolution is beneficial beyond a static skill library. When the skill set is frozen, the average accuracy decreases by 3.1\%, and the average number of tool calls rises from 4.83 to 5.54. This result suggests that continued skill refinement during training remains important even after an initial library has been constructed, since the improved policy can in turn extract and consolidate more effective skills from new interaction experience.
(3) \textbf{Reward Design.} The tool-call penalty is important not only for efficiency but also for overall task performance. Removing the tool-call penalty lowers accuracy by 2.6\% and substantially increases tool calls from 4.83 to 6.83, showing that the penalty helps prevent unnecessarily long tool-use chains and stabilizes learning. In contrast, always penalizing tool calls reduces the average number of calls to 3.96, but causes the largest performance drop, down to 26.1\%. This indicates that overly aggressive regularization suppresses useful exploration and discourages necessary tool use. Overall, the full reward design achieves the best trade-off between task accuracy and tool-use efficiency.

\begin{figure*}[htbp]
\centering
\includegraphics[width=\textwidth]{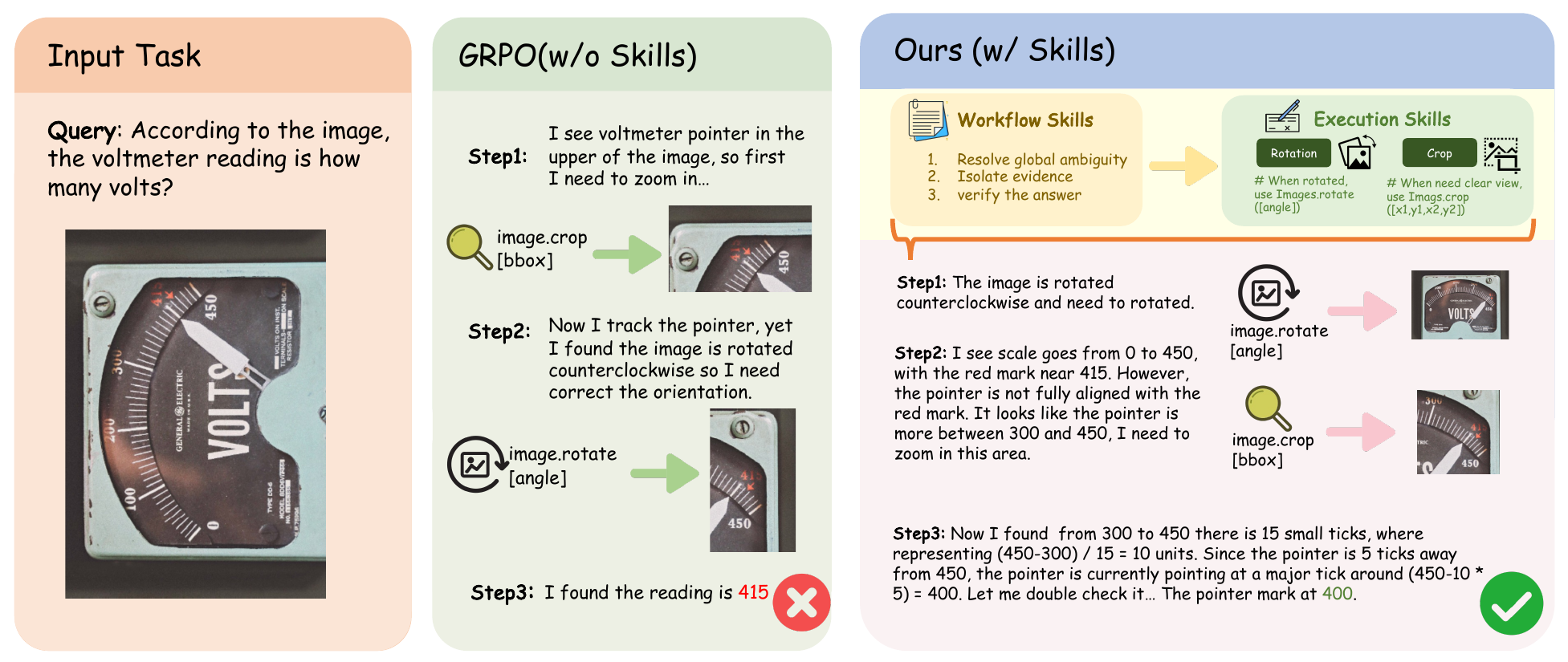}
\caption{Comparison of Reasoning Trajectories on a Multimodal Instrument-Reading Task with and without Retrieved Skills.Without retrieved skills, the agent misuses tools and makes an incorrect local reading. With retrieved skills, it recalls a high-level workflow and execution operations in the right order.}
\label{fig:case_study_skills}
\end{figure*}

\subsection{Analysis}
We provide a detailed analysis of the evolution dynamics for skills. For more analysis, please refer to the supplementary.

\begin{figure}[htbp]
\vspace{-1em}
\centering
\includegraphics[width=\linewidth]{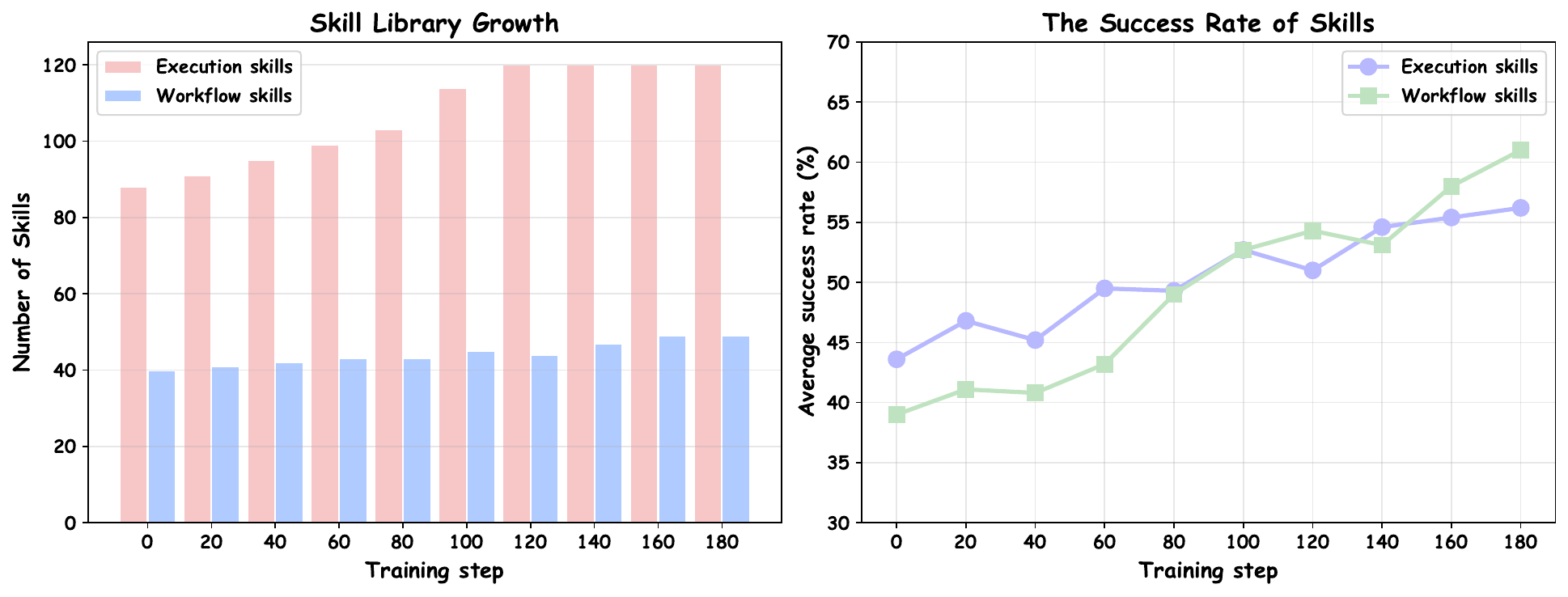}
\caption{Skill library evolution throughout training.}
\label{fig:skill_evolution}
\vspace{-1.5em}
\end{figure}

\noindent \textbf{Skill Library Evolution.}
We monitor the number of workflow skills, the number of execution skills, and the average empirical success rate of execution skills, denoted by $\mathrm{SR}(r)$, during the skill-policy co-evolution stage. As shown in Figure~\ref{fig:skill_evolution}, during the early stage of training, the number of execution skills continues to increase over training, while their average empirical success rate rises. This suggests that policy improvement not only induces the acquisition of new execution skills, but also helps refine existing ones through merging and updating. After around 100 training steps, the library size no longer grows noticeably, as it has reached the predefined capacity limit. Skill evolution is reflected primarily in consolidation and refinement of existing entries rather than further expansion of the library. In contrast, the number of workflow skills remains relatively stable throughout training, with only modest variation over time. This reflects the different functional roles of the two skill levels. Execution skills represent local visual operations, and thus continue to expand as improved policies encounter and solve a broader range of local interaction patterns. Workflow skills, in comparison, represent abstract workflows and tool orchestration strategies, so their evolution mainly takes the form of merging and updating existing structures rather than increasing their count. These observations are consistent with the proposed dynamics of skill--policy co-evolution, in which the skill library provides increasingly effective guidance for policy learning, and the improved policy in turn induces richer and more reliable hierarchical skills from ongoing interaction experience.

\begin{figure}[htbp]
\centering
\includegraphics[width=\linewidth]{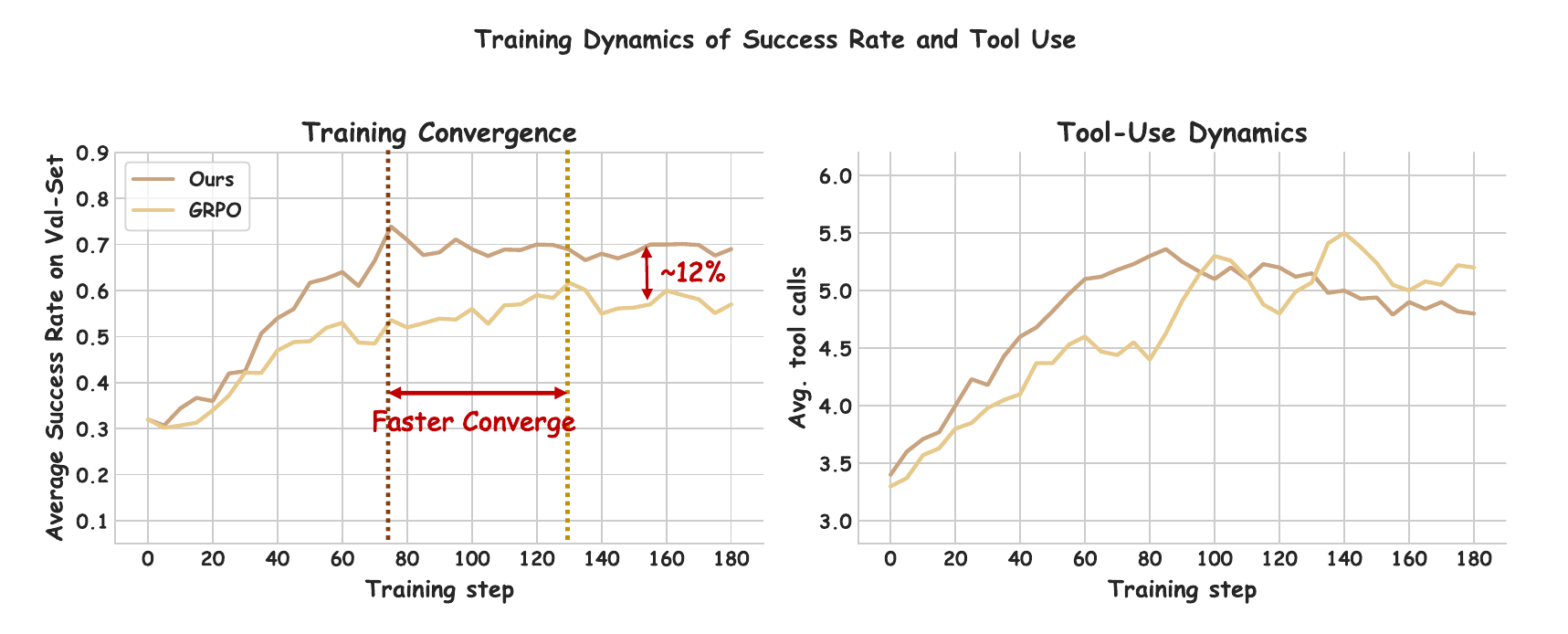}
\caption{Comparison of convergence and tool-use dynamics between SPyCE and GRPO. 
\textbf{Left:} Reward convergence over training. 
\textbf{Right:} Average number of tool calls over training.}
\vspace{-2em}
\label{fig:convergence_tooluse_dynamicsf}
\vspace{-1em}
\end{figure}

\noindent \textbf{Training Convergence and Tool-Use Behavior.}
We compare our method with GRPO in terms of both convergence behavior and tool-use dynamics. As shown in Figure~\ref{fig:convergence_tooluse_dynamicsf}, our method achieves substantially faster improvement in success rate on valitaion set during the early stage of training, with a clear performance gap already emerging after the first few updates. It also converges to a markedly higher success rate, whereas GRPO improves more gradually and remains at a lower, less stable plateau. The difference in tool-use behavior is equally informative. For our method, the number of tool calls increases in the early phase, indicating active exploration of potentially useful operations, but then gradually decreases as training proceeds. This suggests that skill-policy co-evolution does not simply encourage more tool use; rather, it helps the agent convert early exploration into more selective and efficient tool invocation. In contrast, GRPO exhibits a noisier trajectory: tool calls continue to fluctuate at a relatively high level even after success-rate improvement has largely slowed down. Taken together, these results suggest that our framework improves not only final performance but also training efficiency by enabling the policy to discover useful tools early and then compress them into a more disciplined decision strategy.

\noindent \textbf{Case Study.} Figure~\ref{fig:case_study_skills} presents an illustrative comparison between a policy without retrieved skills (only GRPO) and our method with retrieved skills. Without skills, the agent applies tools in a generic and poorly ordered manner: it first crops a local region before correcting the global orientation, and then directly infers the reading from incomplete local evidence, which leads to an incorrect answer. In contrast, with retrieved workflow and execution skills, our method first follows a high-level workflow that resolves the global ambiguity, isolates the relevant evidence, and then verifies the final reading. This workflow is instantiated by targeted execution operations including rotation correction and local crop-based inspection, enabling the agent to focus on the informative scale region and produce the correct answer more reliably.

%% file: 5_conclusion.tex
\section{Conclusion}
In this paper, we propose a skill-policy co-evolution framework for multimodal agents that think with images, which jointly optimizes  policies and a hierarchical skill library during reinforcement learning. By distilling successful trajectories into reusable low-level execution skills and high-level workflow skills, our method enables skills and policy to co-evolve over time. Experiments on three categories of benchmarks show that our method consistently outperforms strong baselines. Further ablations confirm that both the hierarchical skill design and the co-evolution mechanism are critical to the final gains. The analyses further highlight the notable skill evolution dynamics exhibited by our method. Overall, our results suggest that jointly evolving skills and policy provides an effective framework for building more capable and efficient multimodal agents.